# A More Human Way to Play Computer Chess

Kieran Greer, Distributed Computing Systems, Belfast, UK.
http://distributedcomputingsystems.co.uk

Version 1.1

***Abstract***— This paper suggests a forward-pruning technique for computer chess that uses 'Move Tables', which are like Transposition Tables, but for moves not positions. They use an efficient memory structure and has put the design into the context of long and short-term memories. The long-term memory updates a play path with weight reinforcement, while the short-term memory can be immediately added or removed. With this, 'long branches' can play a short path, before returning to a full search at the resulting leaf nodes. Re-using an earlier search path allows the tree to be forward-pruned, which is known to be dangerous, because it removes part of the search process. Additional checks are therefore made and moves can even be re-added when the search result is unsatisfactory. Automatic feature analysis is now central to the algorithm, where key squares and related squares can be generated automatically and used to guide the search process. Using this analysis, if a search result is inferior, it can re-insert un-played moves that cover these key squares only. On the tactical side, a type of move that the forward-pruning will fail on is recognised and a pattern-based solution to that problem is suggested. This has completed the theory of an earlier paper and resulted in a more human-like approach to searching for a chess move. Tests demonstrate that the obvious blunders associated with forward pruning are no longer present and that it can compete at the top level with regard to playing strength.

## 1 Introduction

This paper suggests a forward-pruning technique for computer chess that uses 'Move Tables', which are like Transposition Tables, but for moves not positions. The idea of dynamic move chains has been described in a preceding paper [13]. It incorporates the idea of forward-pruning the search tree, which is known to be dangerous, because it removes part of the search process. As the search heuristic has limited knowledge, if part of the tree is omitted, any information that it contains would also be lost. Because computer chess programs are mainly statistical, they do not typically contain enough knowledge about any single position in their heuristic. The program must therefore evaluate many more positions to reveal potential end results, through a more exhaustive search. Even if evolutionary methods or machine learning is used to improve the evaluations [11], the understanding level compared to a human is still very shallow, although AI will continue to improve this aspect (for example, [17]). Therefore, if the knowledge is going to be limited, a similarly limited technique that can forward-prune that knowledge would be useful. This paper suggests such a technique, where the justification is the integrity between positions along short search paths, makes it more likely that an earlier result still applies.

The intention of dynamic move chains is to store sequences of moves, rather than positions. The move sequences can then be tried in any relevant position, with the expectation that if the move sequence is legal, there is a chance that the saved evaluation of it will be sufficiently accurate. This means that a move sequence that gets stored has a high probability of being relevant to the next position as well, and so on. If you like, dynamic move sequences are more like rules to be applied in a related position. The sequences are stored in lists called chains, that might spawn tree-like structures, but so far in practice, are not very deep structures. The term 'chain' is used to try to distinguish from the more common term 'sequence' that is also used in the paper. If thinking about the human process, then some threats are immediate, but we also store all of the events into a longer-term memory that is built-up over time. When we are faced with a similar situation, we can still retrieve from the long-term memory, helpful advice about what to do. The move chains can also be used for the longer-term memory, where the stored units are move paths with evaluations, and the database tables would be indexed by the first moves.

To support the forward-pruning technique for example, a minimum beam width can guarantee that *x* moves are searched, broadening the minimum search window. The main danger is with tactical moves and a particular type of tactical move has been recognised that is hidden and so it cannot be solved without a search. This paper suggests a solution to that problem, by creating a board representation that may be able to recognise this type of move without a search. Even if all of the tactics can be accounted for, the search may return inferior results by omitting some moves. When this is the case, it is therefore necessary to re-insert some moves that target key squares. A plan or strategy module has therefore been added that can create key features simply from relative piece positions and this can be used to suggest key moves. An idea of 'long branches' is also introduced. They would forward prune only part of the search, for a few ply, before returning to a full search, or quiescence. While the ChessMaps program is not particularly strong, it now incorporates these new ideas and if the program can avoid the obvious blunders, then the forward-pruning will be considered to be successful. Time is another problem however and so the current algorithm is unlikely to be successful in competition, but examples against a top chess program show that it can compete at that level with regard to playing strength.

The rest of the paper is structured as follows: section 2 notes some other chess methods that would be related to the work. Section 3 gives a recap of the dynamic move chains, while section 4 describes the new memory tables idea. Section 5 describes other new work in the area of strategy and automatic feature analysis, while section 6 outlines the whole new search algorithm. Section 7 gives some test results and section 8 ends the paper with some conclusions on the work.

## 2 Related Work

This paper follows on from an earlier paper [13] that introduced the analysis idea, with automatic feature and move path selection, where the move sequences were described as tactical chunks. If the sequence (tactic) is possible in any position, then try it. It also replaces [12] that originally described the dynamic move tables but with some errors. Related research would still be the original Chessmaps Heuristic paper [14], the Killer Heuristic [16][5], Null-Move Heuristics [2][22], the History Heuristic or Transposition Tables [19]. The Chessmaps Heuristic is a move-ordering algorithm that uses square control as a central feature, to distinguish between safe and unsafe moves, and also good or bad regions on the board. The idea is extended further in this paper to include more specific key squares. The Killer Heuristic is a type of immediate result. It can store a killer or best move for each level of the search tree, based on the idea that in many positions, the same move will be the best after any move by the opponent. This can be compared with the short-term memory (section 4.2) that can store the best move for each square on the board, and can be used and updated in a similar manner. Another example of storing moves instead of positions is the Countermove Heuristic [21]. It is based on the observation that in many different positions a particular move is best responded to by some particular opposing move. An economic storage structure can be used to store the moves and also the counter moves. It was shown to reduce the search by as much as 20 – 50%. This could then be compared with paths in Move Tables (section 4), but the tables are used to continue one's own search and not necessarily to counter the opponent's. It is interesting that the forward-pruning search reduction quoted in section 7.3 is also about this amount. While the cutting-edge research is more about evolutionary algorithms and statistical processes, these ideas may be older in nature and knowledge-based, but they can still be generalised.

The History Heuristic uses a similar type of compact storage structure to the Move Tables and Transposition Tables are a position-indexed version of the move-indexed Move Tables. Singular Extensions are another option that has been tried [3] and relate to move paths. They are also domain independent and based on probability, but it was found that they were too computationally expensive, although, variations of it may have been included in top chess programs. Null or Zero-Move heuristics [2][22] are mentioned because they could be useful when trying to formulate plans. With these, a player can make more than 1 move in a row, or the opponent makes a null move. If the free move does not improve the position, then the position is not likely to change much. They are not used in the current ChessMaps program, but if the program were to try to move pieces to particular areas without considering too much every tactic in the position, then allowing some free moves could become important.

The Monte Carlo search [8] was recently used for the AlphaGo program [23] that defeated the leading world Go player and it has also been used to easily win at chess. It also performs selective search and judging from its results, shows what the leading methods are at the moment. The program selects which branches or lines of play to expand further, through statistics, but still performs a full search over what it selects. The advantage of the selective search is that it can deal with almost any problem, including imperfect information and very large branching factors. It may require large search trees to be effective, however and not being able to evaluate intermediate moves is at least a restriction. There are lots of other more recent search algorithms and some have been written about in earlier papers, [13] for example. The 'Method of Analogies' [1] has been suggested and was implemented as part of the Deep Blue program [15]. In the summary they state that:

> '… a game-playing algorithm often inspects the same thing many times over. A human, having studied a situation once, will in the future draw conclusions by the use of analogy. But, it often happens that seemingly insignificant changes in the position alter the course of the game and lead to substantially different outcomes.'

Therefore, specific features and knowledge are required to determine these differences. Or alternatively, lower the specificity level of the knowledge and allow it to prune the search process a bit less. The new research in [17] could also be considered.

## 2.1 Botvinnik's Computer Chess Theory

The former World Champion Botvinnik wrote a computer chess book [7] and program (Pioneer) when the chess programs were built on the theories of logic and knowledge. As computer speeds improved, the statistical brute-force approaches took over and are now standard. The expert logician suggested in his book, to measure how many moves it might take for a piece to move from where it is to where it should be. The chess program is also described in [6], where the following quote from 'The Tale of a Small Tree' chapter sums up some of the relevant philosophy:

> 'Thus, during a game a player analyses the movements of a limited number of pieces on a limited part of the board, and analyses the movements of only those pieces which come directly into collision, and only on those squares where collisions are possible. In other words, he examines only those pieces which interact with the enemy pieces, and only those squares where this interaction is possible. But how can one check that these pieces and these squares have been correctly chosen? …'

This requires specific knowledge about the position and is a bit like narrow sequences that need to be incorporated into a main search process, or preferably, can be seen without the search process. This more specific knowledge represents a plan in the position, but as current programs lack that level of knowledge, a simpler plan devised from more low-level knowledge would be a start. With relation to writing a game-playing program, the following blog [18] is interesting. It actually quantifies the reason why forward-pruning should not work, which is helpful for making comparisons with. It notes that even an evaluation function that is 99% accurate will blunder once in at least a third of its games. Therefore, as the ‘perfect’ evaluation function is not possible, a forward-pruning process needs to be fortified with other types of check, requiring a plan that can intelligently select moves. The ChessMaps Heuristic is quite well setup for this type of plan because it calculates move influence, or the squares a piece can move to after it has moved and so looking one move ahead can be automatic. Therefore, what is required is to recognise the board regions or squares that a plan would ask pieces to move to. That can be as part of square control, but more accurate are the key squares that are described in section 5.

## 3 Dynamic Move Chains

A dynamic move chain [13] is a small move sequence generated from an earlier part of the search. It can be significant because it resulted from a cut-off in the search process. A cut-off is generated when the current evaluation refutes a particular position and means that further search of the position is not required. When a significant result is recognised, the search path can be stored with the result, with the intention of re-using it later instead of a new full search. Even very small changes in a position can produce very different evaluations however and so it is not possible to be 100% accurate by re-playing a move sequence. The move sequence must be re-evaluated again, in any position that it is used, but the stored evaluation can still indicate what move sequences to try. If the move is found to be unreliable, then the evaluation is updated and a more extensive search needs to be made. So, the trade-off is between a small amount of forward pruning, with the risk of a more extensive re-search later, versus a lighter full α-β search every time.

An algorithm for the dynamic move chains is provided in [13] and the similar dynamic move tables also needs to be integrated. The move chains are more temporary and immediate. They are created and updated during the course of the game, but would not be saved. The dynamic tables can store more aggregated information over the course of a game and the results can be saved to files. For any future game, the tables can be read in again, updated and re-saved. In that case they represent a more permanent record of the chess program and even its evolution. This is all described in the next section.

## 4 Memory Tables

The computer program now incorporates dynamic move tables, or memory tables and a shallow feature analysis into the search process. The paper [12] indicates that it is possible to read move sequences from game scores, probably Grandmaster games and play them directly in a search process. While this can work, it is not part of the current program. Including this would mask the effect of the current heuristic set that are more in tune with what the program evaluates for itself. Game scores are read and stored however, to recognise common features for popular moves (section 5.1). The move tables that are used then relate to short or long-term memory and are generated entirely by the program itself.

### 4.1 Dynamic Table Structure

Move Tables perform a similar operation to Transposition Tables [20] and so Transposition Tables are not included in the program, even though they would certainly improve the overall playing strength. The 'Move' title helps to distinguish between indexing on moves or indexing on positions. They have some advantages and some disadvantages over Transposition Tables. For one thing, they can be encapsulated in an array that is the size of a board. They can also be stored for each piece type, for both colours, requiring a total of 12 tables but this is still a relatively small amount of memory compared to Transposition Tables. The tables are then enhanced by linking squares across them that represent valid move paths. The move Nf3, for example, can then belong to more than 1 valid move path. Figure 1 is an example of what the move tables might look like. The large boxes represent the whole board for each piece type and can literally be only 64 elements in size. The smaller squares represent the move sequence that has just been added, resulting in incrementing those related weight values only.

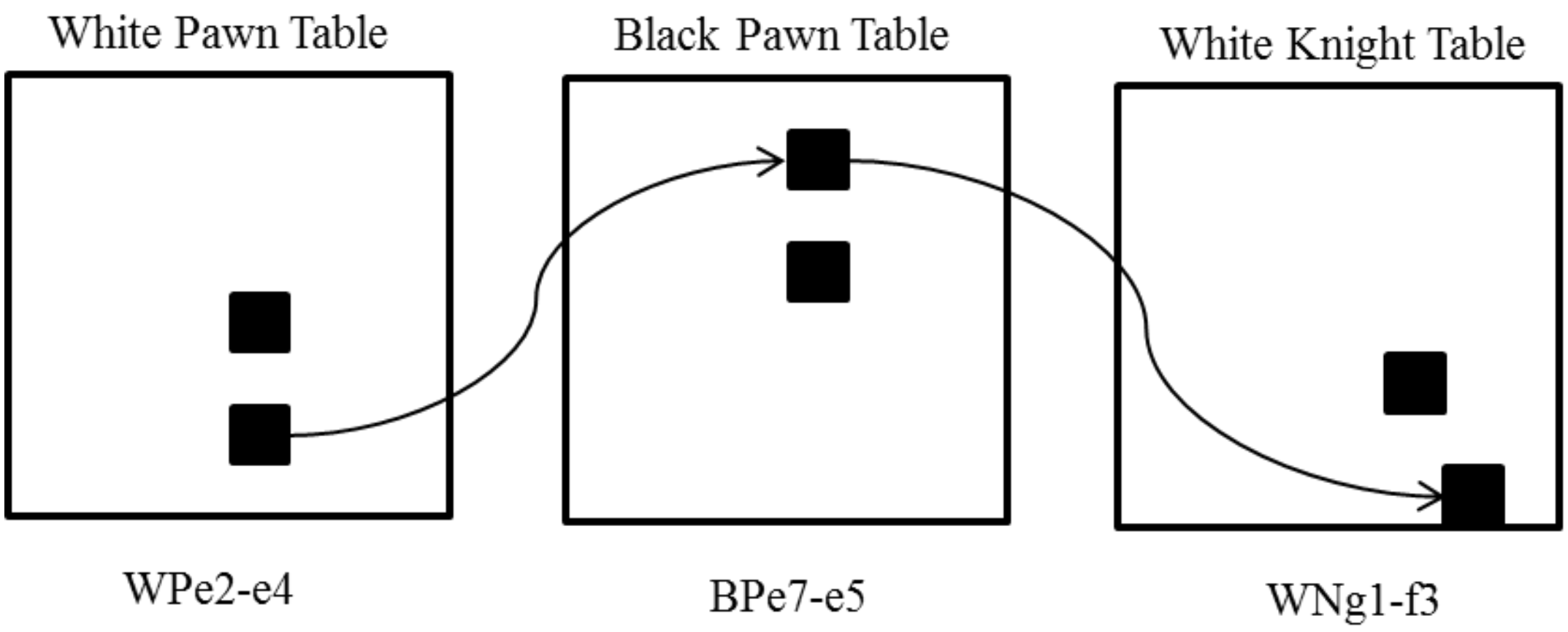

Figure 1. Example of Move Tables and Links between them.

Consider, for example, a cut-off that has resulted in a move sequence such as: WPe2-e4, BPe7-e5, WNg1-f3: Both the 'from' and 'to' move squares can be updated, where the related weight/evaluation value can also be updated. If an evaluation is found to be unreliable however, the table move sequence does not have to be removed immediately, but can have a negative update, until it is eventually removed. There can also be links between the tables, where the move WPe2-e4 is stored in the e2 square that then links to Black's Pawn's table, to the e7 square. That then links to White's Knight's table, to the g1 square, and so on. Any move sequence that is returned can first be checked for legality in the position, before a reliability check. An optimised minimax search can typically have a branching factor that is the square root of all possible moves, but the Move Tables might return more possibilities than this. They should therefore be filtered further, to keep only the top 'x' number, but this can be set to anything and so can be tailored to the particular program that is using it.

### 4.2 Short and Long-Term Memory Tables

Both the short and long-term memory can be built from these move tables but they can then be used differently. The move path is saved under the square that the first piece to move is on and then links from one move to another in the different tables. The short-term memory can use a path of 1 (or more) ply, as in Figure 1. This would be a separate structure to the long-term memory and any entry can be added or removed immediately. In effect, this is a type of Killer Heuristic that stores the best moves over the whole board.

The long-term memory uses a more permanent version of the tables, when 3-ply move sequences are transferred to the long-term table structure. In this case, when a cut-off occurs, the base move is found in the structure and the related path retrieved. The weight and evaluation related to that path can then be updated. The move path itself would not be considered again unless it is both legal and has a reliable weight value. Re-saving the tables in a database is possible and this can store new knowledge generated from each new game. It does not result in very large table sizes however, because there is more weight reduction than reinforcement and it would be found that common move sequences will occur in typical positions. After move sequences are added, it is actually more common for the sequences to be retrieved and have their weights and evaluations updated, than for new sequences to be added. This reaffirms the idea that there is a lot of integrity in a chess position and that it conforms to a restricted set of rules. Therefore after a while, each game might only add a few KBs of moves to the database.

### 4.3 Best Squares and Important Squares

These move tables then produce a pattern of what the best moves for each piece type is over the course of the game or search phase. Figure 2 and Figure 3, for example show some stats from the search analysis that could be useful for knowledge-based reasoning. Each piece can be considered individually, but typically, the results might be aggregated together. If trying to derive some form of knowledge from the move tables, then looking at the 'best' squares only might not be the most beneficial. For example, Figure 2 shows the Queen and Bishop move tables for the indicated position, with only the best positive weights displayed. It clearly agrees with that position, where the best moves for White indicate to attack Black's weakened King's side. The weighted squares however, only relate to two specific moves, which would make any derived knowledge a bit sparse. If there are several moves from a square, for example, then if one of them is the best, the others will have their weight values decremented and might become very small, or they might mutually decrement each other and so frequently considered moves can eventually be assigned negative values.

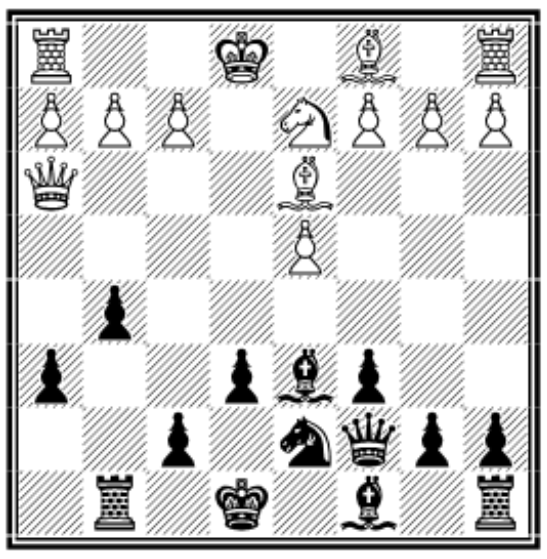

Board Position

White Queen (best) Move Table

| | | | | | | | |
|---|---|---|---|---|---|---|---|
| 0 | 0 | 0 | 0 | 0 | 0 | 0 | 0 |
| 0 | 0 | 0 | 0 | 0 | 0 | 0 | 0 |
| **0.5** | 0 | 0 | 0 | 0 | 0 | 0 | 0 |
| **0** | 0 | 0 | 0 | 0 | 0 | 0 | 0 |
| **0** | 0 | 0 | 0 | 0 | 0 | 0 | 0 |
| **0.5** | 0 | 0 | 0 | 0 | 0 | 0 | 0 |
| 0 | 0 | 0 | 0 | 0 | 0 | 0 | 0 |
| 0 | 0 | 0 | 0 | 0 | 0 | 0 | 0 |

White Bishop (best) Move Table

| | | | | | | | |
|---|---|---|---|---|---|---|---|
| 0 | 0 | 0 | 0 | 0 | 0 | 0 | 0 |
| 0 | 0 | 0 | 0 | 0 | 0 | 0 | 0 |
| 0 | 0 | 0 | 0 | **0.4** | 0 | 0 | 0 |
| 0 | 0 | 0 | **0** | 0 | 0 | 0 | 0 |
| 0 | 0 | **0** | 0 | 0 | 0 | 0 | 0 |
| 0 | **0.4** | 0 | 0 | 0 | 0 | 0 | 0 |
| 0 | 0 | 0 | 0 | 0 | 0 | 0 | 0 |
| 0 | 0 | 0 | 0 | 0 | 0 | 0 | 0 |

Figure 2. Queen and Bishop Move Tables for the shown position.

As an alternative, Figure 3 shows the accumulated weight values for all of the stored moves, for the White Queen. Note that the accumulated values have made every square negative, apart from the squares with the value of 0. These are squares that no moves were stored for and so were not part of the search, whereas, even if the values are negative, those squares have been very actively considered as part of the search. It is also interesting that the squares with the largest negative values relate to the best move for the Queen that is shown in Figure 2.

| | | | | | | | |
|---|---|---|---|---|---|---|---|
| 0 | 0 | 0 | 0 | 0 | 0 | 0 | 0 |
| -246 | 0 | 0 | -65 | 0 | 0 | 0 | 0 |
| **-802** | -213 | -459 | -521 | 0 | 0 | 0 | 0 |
| **-335** | -317 | -98 | 0 | -62 | 0 | 0 | 0 |
| **-739** | -202 | -82 | -47 | 0 | -17 | 0 | -64 |
| **-975** | -61 | -259 | -195 | 0 | -17 | 0 | 0 |
| -36 | -55 | -93 | 0 | 0 | -63 | 0 | 0 |
| -60 | -81 | 0 | 0 | 0 | 0 | 0 | -16 |

White Queen (accum) Move Table

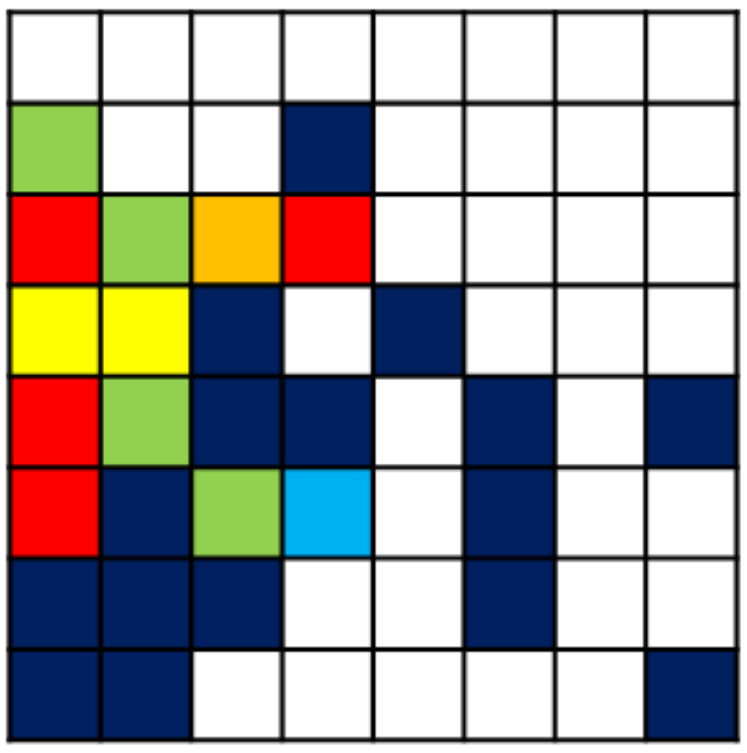

Relative Importance Map View

Figure 3. Accumulated weight strengths and relative importance map.

There are therefore many considerations about best values and one idea might be to take the absolute value of the accumulated weight values to represent a square's 'importance' for a piece and the best weight value(s) to represent the strongest square(s). An important square can then possibly be given priority as part of a plan, for example, as it indicates active lines. To illustrate this, the second diagram of Figure 3 has coloured in the table, using the absolute weight values only. The most important squares are coloured red; then orange, yellow, green, light blue and dark blue is the least important. The white squares should maybe not be considered at all.

### 4.4 Evaluation Function Series

When returning an evaluation for a chess move from a table, there is then the problem of what value to return, as it can be different for a different position. Currently, the program stores the last 10 evaluations in order, with most recent to least, played in increasingly different positions from the current position. So, the most recent evaluation must have the most influence. If a time factor is not considered, then one option would be something like (E1 + E2/2 + E3/3), and so on, where E1 is the most recent evaluation, E2 is the second most recent, and so on. But with this, the importance of the most recent evaluation score 'E1' is quickly lost, as the other scores are included. Therefore, an alternative to this might be:

$$(E1 + (E1+E2)/2 + (E1+E2+E3)/3 + \ldots + (E1+E2+\ldots+En)/n) / n$$

This is a mathematical series and with it, the influence of E1 decreases more gradually throughout the whole series, followed by E2, and so on. So, this may be a more attractive function for smoothing out the relative score's importance. It might be interesting to see this series as a line of neurons in the brain, where the first firing neuron is the most important, or recent in this case and the last firing neuron is the least important.

## 5 Strategy and Feature Analysis

The feature analysis of this section completes the theory of the paper [13] and allows the program to suggest key squares that in turn can suggest key moves. While square control can suggest areas to influence, key squares can be more specific. This can include popular squares that are from general knowledge, or key squares that are from the search. As described in the following sections, popular squares can be derived from a database created from game scores, while key squares can be derived from feature analysis during the game.

### 5.1 Features from a Database

It is possible to read in game scores and generate a database of move frequencies from typical positions. The database is indexed on the move itself, and then stored with that is a frequency for each piece that may occur on each square. A decision has been made to use move paths of 3 ply for long-term storage. This requires some consistency in the position but can still be played in a game. Also, a move sequence was added only if it occurred at least 3 times, when it would then have occurred in different positions and so probably be consistent with some inherent positional structure. These knowledge-based move sequences were saved to a database that ended up being 22M in size, from maybe 2-3 million different moves.

### 5.2 Popular and Key Squares

As part of the α-β search, moves to popular squares can be identified from the move frequency database just described. A piece frequency board is retrieved with the most popular pieces based on some stats. If any moves in the current list match with these pieces, then add them to the search. Influence can mean both moving to the square or attacking it after a move. If tactical moves are always considered in the search, then popular squares would prune the 'safe other' or non-capture moves. As part of the ChessMaps Heuristic [14], this category is usually ordered by the neural network, but the database can be much more specific, with the neural network producing a secondary ordering.

While popular squares are a basic statistical count, another analysis finds features by comparing two critical positions. The first is the root position in the search sequence and the second is a subsequent position along the search path, where the evaluation score changed markedly, possibly outside of some quiescence value. Differences in the positions can be recognised for both sides and can be stored as both strong and weak features for either side. Typically, if a piece is present or missing on one of the two boards, then it is a difference. These relate to squares that need to be covered in the current position and the comparison can be automatic, as the earlier paper suggested [13]. This process is currently carried out when a position receives an inferior evaluation score, but there are still some ‘safe other’ moves that have not been tried. After identifying these feature squares, any moves that influence them can be re-added and then searched over.

### 5.3 Hidden Tactical Moves

This section outlines another type of move that is not recognised by the ChessMaps Heuristic and would probably not be added through the search algorithm. It is tactical in nature but slightly hidden and so traditionally, a search would be required to find it. If the position needs to be searched however, the forward-pruning becomes ineffective and so the move type needs to be recognised from patterns in the board position. It is also important to try to solve this problem in a domain independent way, so that the method can be transferred to other domains. The move category in question can be described in the chess position of Figure 4. This position was realised through a game between the ChessMaps program and the Arena program running the SOS 5.1 chess engine [4].

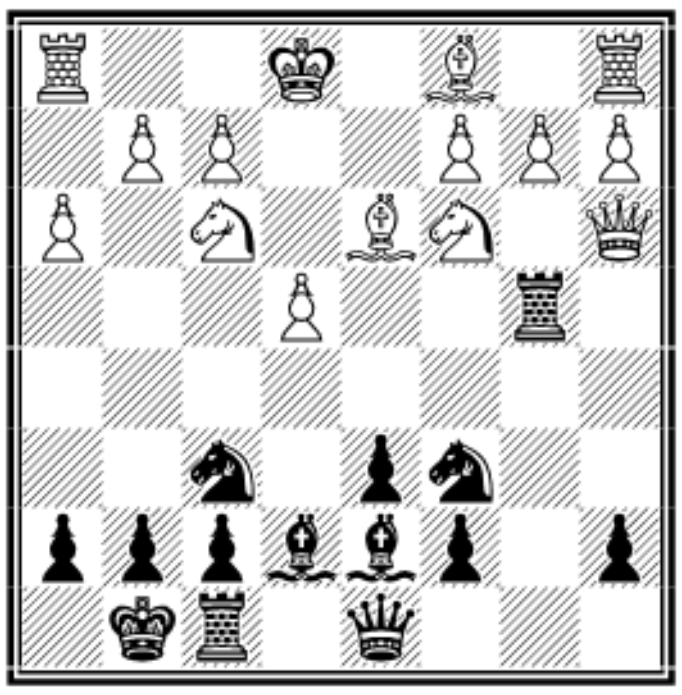

Figure 4. There is an indirect attack of the Black Bishop on the White Queen that can be realised through some forcing moves.

The last move for White was to move the Queen from a4 to a3, which looks safe, but now there is an indirect attack on the Queen from the Black Bishop on e7. The diagonal is blocked by 2 other pieces, but they can move out of the way with forcing moves and thereby expose the Queen to the attack. The move sequence would be something like … d5, exd5 Re4+, Nxe4 Bxa3 with a win. The problem is that the forward-pruning algorithm does not consider this move sequence because the intermediate moves are all unsafe and so the search would miss this sequence.

It is however, possible to recognise this scenario without a search and in a general way. All that is required is to know the relative value of the pieces and how they move. When the move ordering is generated, a new feature board can be created that stores for any long-range piece, the squares along any row/column or diagonal that it is aligned with. For that piece, it also stores any other pieces on those squares. In the position of Figure 4, for example, the new structure would store information for the Black Bishop on e7, because it is a long-range piece. Looking at that Bishop shows that it is aligned with the White Queen on a3, but also that there are two other Black pieces in the way. The worst-case scenario is if they are both lost but then the capture is possible. Adding-up the values of the pieces on the diagonal shows that even if the two blocking pieces are lost, the threat to the Queen will still result in a material gain. Therefore, a decision can be made to consider all moves (both safe and unsafe) of those two pieces and this type of tactic can be included in the search. Granted these are extra moves that should be forward-pruned, but the scenario is a rare one and so in this special case.

## 6 New Search Algorithm

With the techniques described in the preceding sections, it is now possible to suggest a new search algorithm for computer chess. With the success of current algorithms, it is unlikely to be more successful, but it is based more on the steps a human takes and may be of interest for that reason. The move ordering for the search process still starts with the ChessMaps Heuristic. This is able to recognise different types of move that are as follows:

1. Safe capture moves, Safe forced moves, Safe forcing moves.
2. Safe other moves.
3. Unsafe capture moves, Unsafe forced moves, Unsafe forcing moves.
4. Unsafe other moves.

One division is to split the moves into safe and unsafe. Unsafe moves would lead to loss of material on the square the piece is moved to, while safe moves would not. It is also possible to recognise capture, forced and forcing moves. Forced moves would be ones where the piece would have to move because it could be captured with a loss of material. Forcing moves are moves that force the opponent to move a piece because it could then be captured with a gain of material. Capture moves capture a piece and other moves move a piece to an empty square, without an immediate threat.

### 6.1 Tactical Moves

The tactical moves always need to be considered and after the full α-β search there is a quiescence search of safe tactical moves only. Therefore, the first moves that are considered in the full α-β search are the safe captures, forcing and forced moves. It is also useful to add the unsafe capture moves to the search list, because search sequences can include pinned pieces that are not directly obvious.

### 6.2 Position Analysis

Because of the forward-pruning, the position needs to be analysed further, to realise critical features. This is currently more expensive than using a lightweight full search process, but that is the direction for the theory of this paper.

#### 6.2.1 Exposing Moves

The hidden threat move type described in section 5.3 should now be added, as well as moves that block a check, but this could be a problem specific to the ChessMaps program.

### 6.2.2 Feature Moves

After that, the safe other moves need to be considered and this is where the forward pruning takes place. In the ChessMaps Heuristic, this move category is ordered using a neural network, that recognises square control. So that this is still useful, the first 4 or more moves ordered by the neural network are added. Then a set of moves that tally with the strategy database of section 5.1 is added. A list of piece frequencies is retrieved from the database based on popular statistical counts. The most popular pieces for each square are then calculated[1] over all of the retrieved frequencies. The current move list is then traversed and if any move agrees with the combined board piece positions, it gets included. Typically, the square on the feature board should store a piece that is of the same type and colour as for the considered move, or the considered move can influence (attack) the square.

### 6.2.3 Re-Add Feature Moves to the Search

Popular moves are automatically considered as part of the full search. After that filtering phase, there are still some safe moves and all of the unsafe forced, forcing or other moves that are not included and so forward-pruning will always remove these. If the search returns an inferior score, then these moves are considered again. So when a poor score is returned, another analysis phase looks at the search path and the related positions. It determines the strong and weak squares for either side and tries to cover those key squares with any of the remaining moves. If any remaining move can cover any of the feature analysis squares, they are re-added and searched over. The intention is not to try every move, but to re-add moves that target specific problems in the position.

## 6.3 Memory Moves

The memory tables are used, both as part of the full search and as a separate partial search. Popular squares have already added moves to the full search and the short-term memory moves are also added to it. As the long-term table contain 3-ply paths, if a path is considered to be reliable, it can be played as the search result from the move, followed by a quiescence search at the end. So the long-term memory moves are played during a separate search stage and the result is then compared with the full search. The long-term memory moves can be filtered first, to return the move path with the best evaluation, for example. Then only one path needs to be played and re-evaluated from the current position.

[1] Calculating the frequencies could possibly be more economically done, as a pre-processing operation.

### 6.3.1 Update the Memory Tables

The search process also needs to update the memory tables and they can both be updated at the same time. Because capture moves are always considered, it is decided not to include them in the memory tables and so a check for this is made. Then, when either search phase results in a cut-off, the memory tables are positively updated. For the short-term move chains, the move is added to the tables. For the long-term dynamic table, the related weight value is incremented. If the search results in a worse move, then the tables are negatively updated. For the move chains, the move is removed and for the dynamic tables, the related weight is decremented. For either case, the new evaluation score is added.

## 6.4 Final Move Decision

After the initial searches, the position has a choice of two moves – one from the full search and one from the long-term memory search. If the long-term memory result is better, then a new full search is carried out from the first move only, as a final check. This repeats the whole search process, but from only 1 move. If the result is still poor, then re-adding moves using key squares is considered. After that search, a decision can be made about the best move in the position.

# 7 Testing

The chess-playing program and the test program are both written in the C# .Net language. With an earlier version of the algorithm, two types of test have been carried out and the results are given in the first two test sections. These are consistent with the tests carried out at that time and further tests would not change the numbers significantly. A third test describes the result of playing the algorithm against one of the top chess programs.

### 7.1 Play Against the Original ChessMaps Program

The first is simply some 5-minute games against the original Chessmaps Heuristic, to show that using Move Tables with Dynamic Move Chains does not have a detrimental effect in real play. A random opening for each game would be selected and there was also iterative deepening with a minimal window. The Dynamic Move Chains with Move Tables performed well enough against the standard Chessmaps Heuristic with Transposition Tables. In ten 5-minute games, it won 7-3.

### 7.2 Tests Varying the Long-Term Memory

To see how adding the move tables changes the search size, tests adding 1, 2 and 4 moves to the start of the Move Chains search process was measured. The paper [14] shows how the forward pruning approach of using dynamic move chains by itself, can reduce the search by as much as 99% more than a standard one and so there is a lot of scope for trying to broaden the search process by adding extra knowledge. Table 1 gives the results of the relative search sizes for dynamic move chains with or without the additional long-term move tables. This was measured over a full game containing 98 separate positions to a depth of 5 ply. The results are for the negamax α-β search stage only and show the number of times the search size is less rather than the reduction percentage. Move chains by itself searched 23 times fewer nodes than the broader searches. Note that this result does not measure move quality, only the search reduction.

| **Search Type** | Move Chains | Move Chains + Move tables (1 move) | Move Chains + Move tables (2 moves) | Move Chains + Move tables (4 moves) |
|---|---|---|---|---|
| **Av Nodes** | 545 | 13367 | 12117 | 12550 |

Table 1. Search reduction for Move Chains, or MC plus Move Tables – 1, 2 or 4 moves.

It is interesting that adding 1 single additional move dramatically increases the search size (98%), but adding more than 1 move does not have the same increase effect. So if using the short-term Move Chains by itself, it searches in effect, an average branching factor of 3-4.

Then forcing at least 1 additional full-search move, increases this to 6-7, which is the standard search size. Forcing more than 1 additional move does not have the same effect. So further down the search, where all of the cut-off techniques are still used, the new additions must be compensated in some way, possibly by adding accuracy to the move ordering. Another stat here suggests that move tables are used maybe only 30% of the time, as part of the α-β search, or not used 70% of the time, when the dynamic cut-offs can still occur. But this is for a new table from a single position. If the tables are built up from earlier searches during a game, then the earlier history would make them more relevant.

### 7.3 Play Against a Top Chess Program

The Arena Chess GUI [4] was selected, as well as the default SOS 5.1 chess engine, which is a Grandmaster-strength chess-playing program. The test fixed the search depth to 5-ply for either side. This was done to allow the Arena program to still make good moves and keep the ChessMaps program inside a reasonable time limit. There was in fact, no comparison between the amount of time taken, where Arena would play instantly compared to possibly minutes by the ChessMaps program. The ChessMaps program however is still incomplete in some areas and so a full game that went into the ending would always favour the more professional program. Therefore, the result would be a judgement about general playing strength and number of moves survived before the loss was obvious. The amount of forward-pruning was also measured, as the percentage number of other moves that were not played. Over 20 games, ChessMaps did in fact achieve two clear wins. Tactics are still the problem however, where it missed winning opportunities in 3 other games and maybe scored 3-4 draws as well. Appendix A gives the score for the first won game. With regard to reducing the number of moves considered: a 25-35% reduction would not be uncommon, although it could get as low as 10-15% and increase to 40-50%, for example.

## 8 Conclusions

This paper has added to the forward-pruning technique of earlier papers, through the use of Move Tables that can act in the same way as Transposition Tables, but for moves not positions. They use an efficient memory structure and have put the design into the context of short or long-term memories. So, there are different views of the same knowledge – one is immediate and one is more aggregated. Automatic analysis of the positions is also possible, when key features can be identified and used to select specific moves, along with a more general assessment of square control using the neural network. While the whole search process seems more human-like, the resulting computer program is not particularly strong, but there may be other reasons for that. The code has not been optimised and so while a lot of extra calculations are required, the code could certainly be speeded up. Even a depth of 5 ply for the α-β search at the moment, is probably too slow. The intention of this paper is therefore to demonstrate that the new algorithm can incorporate a number of new techniques that remove obvious blunders from a forward-pruning process.

The feature analysis has advanced the future work theory of the earlier paper [13] and made more explicit where logical plans or more knowledge-based approaches might be tried. Botvinnik's idea of longer range plans is appropriate, because key squares can be covered by a move and not necessarily moved directly onto. So that would be a plan of 2 moves at least and it has probably also led to a new type of move structure that allows for shallow calculations instead of requiring a search. It is also likely that successful results would be transferrable to other types of game or problem as the knowledge is quite general. Indexing on moves can make them inaccurate in some position, but it also stores partial information that can be relevant in a general sense. It may be that positions with missing information can still try legal move sequences that match with certain features, for example. To help with the accuracy that is based on imperfect information, therefore requires the results to be aggregated, so that they represent a range of relevant positions and not one specific position. The results here also suggest that moves need to be re-added, when the first set proves to be too inaccurate, but this is again a human-like process of adjusting from feedback.

## 9 References

[1] Adelson-Velsky, G., Arlazarov, V. and Donskoy, M. (1988). Algorithms for Games, Springer.

[2] Adelson-Velskiy, G.M., Arlazarov, V.L., & Donskoy, M.V. (1975). Some methods of controlling the tree search in chess programs. Artificial Intelligence, Vol. 6, No. 4, pp. 361 - 371.

[3] Anantharaman, T., Campbell, M. and Hsu, F-h. (1988). Singular extensions: Adding Selectivity to Brute-Force Searching. AAAI Spring Symposium, Computer Game Playing, pp. 8-13. Also published in ICCA Journal, Vol. 11, No. 4, republished (1990) in Artificial Intelligence, Vol. 43, No. 1, pp. 99-109. ISSN 0004-3702.

[4] Arena 3.5.1. (2018), web site www.playwitharena.com. (last accessed 8/6/18).

[5] Birmingham, J. and Kent, P. (1977). Tree searching and tree pruning techniques, Advances in Computer Chess, Edinburgh University Press, Ed. Clarke, M.R.B., pp. 89-107.

[6] Botvinnik, M.M. (1981). Selected Games 1967 – 1970, Pergamon Russian Chess Series, Pergamon Press, translated by K.P.Neat.

[7] Botvinnik, M.M. (1970). Computers, Chess, and long range planning, New York: Springer-Verlag.

[8] Browne, C., Powley, E., Whitehouse, D., Lucas, S., Cowling, P., Rohlfshagen, P., Tavener, S., Perez, D., Samothrakis, S. and Colton, S. (2012). A Survey of Monte Carlo Tree Search Methods, IEEE Transactions on Computational Intelligence and AI in Games, Vol. 4, No. 1, pp. 1 - 49.

[9] Chessmaps program, http://distributedcomputingsystems.co.uk/gameplaying.html. (last accessed 30/6/17).

[10]Fürnkranz, J., (1995). Machine Learning in Computer Chess: The Next Generation, ICCA Journal, Vol. 19, No. 3, pp. 147-161.

[11]Fürnkranz, J. (2007). Recent advances in machine learning and game playing. OGAI Journal, Vol. 26, No. 2. Special Issue on Computer Game Playing.

[12]Greer, K. (2015). Dynamic Move Tables and Long Branches with Backtracking in Computer Chess, available on arXiv at http://arxiv.org/abs/1503.04333.

[13]Greer, K. (2013). Tree Pruning for New Search Techniques in Computer Games, Advances in Artificial Intelligence, Vol. 2013, Article ID 357068, 9 pages. doi:10.1155/2013/357068, Hindawi.

[14]Greer, K., (2000). Computer Chess Move Ordering Schemes Using Move Influence, Artificial Intelligence Journal, Vol. 120, No. 2, July, pp. 235-250.

[15]Hsu, F. (2002). Behind Deep Blue: Building the Computer that Defeated the World Chess Champion. Princeton University Press. ISBN 0-691-09065-3.

[16]Huberman, B.J. (1968). A Program to Play Chess End Games, Technical Report, No. CS-106, Ph.D. Thesis, Stanford University, Computer Science Department.

[17]Lai, M. (2015). Giraffe: Using Deep Reinforcement Learning to Play Chess. M.Sc. thesis, Imperial College London, arXiv:1509.01549v1.

[18]Laramee, F.D. (2000). Chess Programming Part III: Move Generation, http://www.gamedev.net/page/resources/_/technical/artificial-intelligence/chess-programming-part-iii-move-generation-r1126. (last accessed 30/6/15).

[19]Schaeffer J., (1989). The History Heuristic and other Alpha-Beta Search enhancements in Practice, IEEE Transactions on Pattern Analysis and Machine Intelligence, Vol. 11, No. 11.

[20]Schaeffer J., and Plaat A., (1996). New Advances in Alpha-Beta Searching, Proceedings of the 25th Computer Science Conference.

[21]Uiterwijk, J.W.H.M. (1992). The Countermove Heuristic, ICCA Journal, Vol. 15, pp. 8 – 15, ISSN 0920-234X.

[22]Winands, M.H., Van den Herik, H.J., Uiterwijk, J.W. and Van der Werf, E.C. (2005). Enhanced forward pruning. Information Sciences, Vol. 175, No. 4, pp. 315-329.

[23]Yustalim, W. (2017). Application of Monte Carlo Search Tree in AlphaGo, http://informatika.stei.itb.ac.id/~rinaldi.munir/Matdis/2016-2017/Makalah2016/Makalah-Matdis-2016-010.pdf.

## Appendix A – Games Score vs Arena

This appendix lists the score for one game played against the Arena Chess Program [4] using the SOS 5.1 chess engine. The playing strength was set to a fixed depth of 5-ply but any amount of time was allowed. The final position was also judged to be a clear win by Arena.

| | Chessmaps | vs Arena |
|---|---|---|
| 1. | e4 | e5 |
| 2. | Nf3 | Nc6 |
| 3. | Bb5 | Nf6 |
| 4. | O-O | Be7 |
| 5. | Re1 | O-O |
| 6. | d3 | d6 |
| 7. | Be3 | Ng4 |
| 8. | Bd2 | Nd4 |
| 9. | Nxd4 | exd4 |
| 10. | h3 | a6 |
| 11. | hxg4 | axb5 |
| 12. | Na3 | Bd7 |
| 13. | c3 | bxc3 |
| 14. | Bxc3 | d5 |
| 15. | Nc2 | Bd6 |
| 16. | exd5 | Qh4 |
| 17. | Ne3 | b4 |
| 18. | Bd4 | Ba4 |
| 19. | Qf3 | Qh2+ |
| 20. | Kf1 | Rae8 |
| 21. | g3 | b3 |
| 22. | axb3 | Bb5 |
| 23. | Ke2 | Qh6 |
| 24. | Rad1 | f6 |
| 25. | Kd2 | Kh8 |
| 26. | Rh1 | Qg6 |
| 27. | Nf5 | h6 |
| 28. | Nxd6 | cxd6 |
| 29. | Rh5 | Rc8 |
| 30. | Qf5 | Qf7 |
| 31. | Ra1 | Rfe8 |
| 32. | Be3 | Bd7 |
| 33. | Qf4 | Qg6 |
| 34. | Rh4 | Kg8 |
| 35. | Qb4 | Re5 |
| 36. | Qxb7 | Re7 |
| 37. | Ra5 | Kf7 |
| 38. | Qb4 | Rc7 |
| 39. | Ra6 | Bc8 |
| 40. | Rxd6 | Bd7 |
| 41. | Rb6 | Re8 |
| 42. | Rb7 | Rxb7 |
| 43. | Qxb7 | Rd8 |
| 44. | Bb6 | Ke8 |
| 45. | Bxd8 winning | |